\pdfoutput=1
\documentclass{article}

\usepackage[preprint]{corl_2026}

\makeatletter
\renewcommand{\@noticestring}{CoRL 2026 Workshop ``Do Robots Need World Models?'', Austin TX, USA.}
\makeatother
\hypersetup{
  pdftitle={Detecting Is Not Choosing: What a World Model Needs to Keep a VLA Policy Safe},
  pdfauthor={Thanish Bolla},
  pdfsubject={CoRL 2026 Workshop: Do Robots Need World Models?}
}

\usepackage{amsmath,amssymb}
\usepackage{graphicx}
\usepackage{booktabs}
\usepackage{multirow}
\usepackage{xcolor}
\usepackage{xspace}
\xspaceaddexceptions{]\%}
\graphicspath{{generated/}{figures/}}

\newcommand{\pizerofive}{$\pi_{0.5}$\xspace}
\newcommand{\pizerofast}{$\pi_0$-FAST\xspace}
\newcommand{\ehalf}{\ensuremath{\epsilon_{1/2}}\xspace}

\IfFileExists{generated/numbers.tex}{%
\newcommand{\nSpPi}{400\xspace}

\newcommand{\fixedSpPi}{46\xspace}

\newcommand{\nLtPi}{200\xspace}
\newcommand{\baseLtPi}{70.5\xspace}

\newcommand{\fixedLtPi}{50\xspace}

\newcommand{\succLtPiLo}{$-$8.0\xspace}
\newcommand{\succLtPiHi}{$-$2.0\xspace}
\newcommand{\nSpFast}{500\xspace}

\newcommand{\gainLtPlus}{25.7\xspace}

\newcommand{\seN}{200\xspace}
\newcommand{\seSafeDisc}{1\xspace}
\newcommand{\seSuccDisc}{0\xspace}
\newcommand{\nRC}{107\xspace}
\newcommand{\nRCall}{150\xspace}

\newcommand{\gainRCClLo}{10.5\xspace}
\newcommand{\gainRCClHi}{52.7\xspace}
\newcommand{\fixedRC}{33\xspace}

\newcommand{\rcCoffeeN}{47\xspace}
\newcommand{\rcCoffeeGain}{51.1\xspace}
\newcommand{\rcCoffeeLo}{36.2\xspace}
\newcommand{\rcCoffeeHi}{66.0\xspace}
\newcommand{\rcCoffeeFix}{24\xspace}

\newcommand{\rcCoffeeSuccBase}{8.5\xspace}
\newcommand{\rcCoffeeSuccGuard}{12.8\xspace}
\newcommand{\rcSinkN}{44\xspace}
\newcommand{\rcSinkGain}{13.6\xspace}
\newcommand{\rcSinkLo}{4.5\xspace}
\newcommand{\rcSinkHi}{25.0\xspace}

\newcommand{\rcSinkSuccBase}{95.5\xspace}

\newcommand{\rcBlenderN}{16\xspace}
\newcommand{\rcBlenderGain}{18.8\xspace}
\newcommand{\rcBlenderLo}{0.0\xspace}
\newcommand{\rcBlenderHi}{37.5\xspace}

\newcommand{\rcBlenderSuccBase}{6.3\xspace}
\newcommand{\rcBlenderSuccGuard}{12.5\xspace}
\newcommand{\gainMin}{11.5\xspace}
\newcommand{\gainMax}{30.8\xspace}
\newcommand{\nGuardRows}{8\xspace}
\newcommand{\brokenMax}{7\xspace}
\newcommand{\devSpPi}{3.6\xspace}
\newcommand{\devLtPi}{4.9\xspace}
\newcommand{\succLtPiAbs}{83.0\xspace}
\newcommand{\succLtPiBase}{88.0\xspace}

\newcommand{\radMedA}{0.53\xspace}

\newcommand{\sigA}{0.45\xspace}
\newcommand{\radMedB}{1.06\xspace}

\newcommand{\sigB}{0.9\xspace}
\newcommand{\radMedC}{3.18\xspace}

\newcommand{\sigC}{2.7\xspace}

\newcommand{\keptSpCdyn}{61\xspace}

\newcommand{\dynMinusStaticSp}{+3.5\xspace}
\newcommand{\dynMinusStaticSpLo}{1.0\xspace}
\newcommand{\dynMinusStaticSpHi}{6.0\xspace}

\newcommand{\keptLtCdyn}{79\xspace}

\newcommand{\halfNSp}{18\xspace}

\newcommand{\whenSpTwo}{0.95\xspace}
\newcommand{\selSpTwo}{0.61\xspace}

\newcommand{\selChanceSp}{0.28\xspace}
\newcommand{\selExactSp}{1.00\xspace}
\newcommand{\halfRadSp}{1.57\xspace}
\newcommand{\halfRadSpLo}{0.31\xspace}
\newcommand{\halfRadSpHi}{3.13\xspace}
\newcommand{\lostSpC}{70\xspace}
\newcommand{\lostSpCdyn}{39\xspace}

\newcommand{\halfNLt}{12\xspace}

\newcommand{\halfRadLt}{3.27\xspace}
\newcommand{\halfRadLtLo}{2.65\xspace}
\newcommand{\halfRadLtHi}{3.73\xspace}
\newcommand{\lostLtC}{34\xspace}
\newcommand{\lostLtCdyn}{21\xspace}

\newcommand{\lvRadTwo}{2.35\xspace}

\newcommand{\lvRadEight}{9.42\xspace}
\newcommand{\physLtFricMinus}{0.69\xspace}
\newcommand{\physLtFricPlus}{0.64\xspace}
\newcommand{\stateLtEventEight}{0.82\xspace}
\newcommand{\evHalfRadSp}{0.21\xspace}
\newcommand{\evHalfRadSpLo}{0.04\xspace}
\newcommand{\evHalfRadSpHi}{1.37\xspace}

\newcommand{\evSpTinyRad}{0.01\xspace}
\newcommand{\evSpTiny}{0.82\xspace}
\newcommand{\evSpTinyLo}{0.73\xspace}
\newcommand{\evSpTinyHi}{0.90\xspace}
\newcommand{\evSpTinyN}{50\xspace}
\newcommand{\evChanceSp}{0.37\xspace}
\newcommand{\evChanceLt}{0.44\xspace}

\newcommand{\lpBelief}{25\xspace}

\newcommand{\lpN}{200\xspace}

\newcommand{\lpVsGaussAbs}{6.5\xspace}
\newcommand{\lpVsGaussLo}{$-$11.0\xspace}
\newcommand{\lpVsGaussHi}{$-$2.5\xspace}
\newcommand{\lpErrByst}{0.56\xspace}
\newcommand{\lpErrBystPn}{3.4\xspace}
\newcommand{\lpErrTarg}{2.85\xspace}
\newcommand{\lpErrTargPn}{19.5\xspace}
\newcommand{\fpTrSpByst}{1.17\xspace}

\newcommand{\fpTrSpTarg}{1.27\xspace}

\newcommand{\fpTrLtTargPn}{468\xspace}

\newcommand{\fpRescueTrSp}{0.31\xspace}
\newcommand{\fpRescueTrSpLo}{0.12\xspace}
\newcommand{\fpRescueTrSpHi}{0.50\xspace}

\newcommand{\fpRescueNSp}{26\xspace}
\newcommand{\fpRescueTrLt}{0.69\xspace}
\newcommand{\fpRescueTrLtLo}{0.57\xspace}
\newcommand{\fpRescueTrLtHi}{0.81\xspace}

\newcommand{\fpRescueNLt}{78\xspace}
\newcommand{\cwPred}{6.5\xspace}

\newcommand{\cwStill}{5.8\xspace}

\newcommand{\cwSel}{0.42\xspace}
\newcommand{\cwSelChance}{0.34\xspace}
\newcommand{\cwSelN}{26\xspace}
\newcommand{\cwSelTrue}{0.65\xspace}
\newcommand{\cwSelSim}{0.81\xspace}
\newcommand{\ltErrSp}{3.5\xspace}
\newcommand{\ltAucSp}{0.46\xspace}

\newcommand{\ltAucLt}{0.45\xspace}

\newcommand{\critWithinMin}{0.551\xspace}
\newcommand{\critWithinMax}{0.560\xspace}
\newcommand{\critWithinBestHi}{0.591\xspace}
\newcommand{\critPooledMin}{0.778\xspace}
\newcommand{\critPooledMax}{0.864\xspace}
\newcommand{\critPooledState}{0.864\xspace}

\newcommand{\critSelP}{0.56\xspace}
\newcommand{\critSelPLo}{0.47\xspace}
\newcommand{\critSelPHi}{0.64\xspace}
\newcommand{\critSelChance}{0.39\xspace}
\newcommand{\critNcontested}{289\xspace}
\newcommand{\critNdanger}{104\xspace}
\newcommand{\critNepisodes}{200\xspace}

\newcommand{\critLossDelta}{$-$0.007\xspace}

\newcommand{\critTrainDec}{46{,}202\xspace}
\newcommand{\critTrainContested}{1{,}641\xspace}
\newcommand{\dynWithinCF}{0.571\xspace}

\newcommand{\dynWithinCFMin}{0.535\xspace}
\newcommand{\dynWithinCFMax}{0.607\xspace}
\newcommand{\dynNcontested}{53\xspace}
\newcommand{\cosmosValDS}{$-$1.0\xspace}
\newcommand{\cosmosValDSLo}{$-$4.5\xspace}
\newcommand{\cosmosValDSHi}{2.0\xspace}

\newcommand{\ceilSp}{43\xspace}
\newcommand{\ceilRatioSp}{0.75\xspace}
\newcommand{\ceilRatioSpLo}{0.65\xspace}
\newcommand{\ceilRatioSpHi}{0.85\xspace}

\newcommand{\ceilLt}{44\xspace}

\newcommand{\restN}{106\xspace}
\newcommand{\restNoise}{6\xspace}
\newcommand{\restZeroVel}{42\xspace}
\newcommand{\restSettle}{46\xspace}
\newcommand{\restHold}{66\xspace}
\newcommand{\restArm}{76\xspace}
\newcommand{\steerOffN}{171\xspace}
\newcommand{\steerOffSamples}{6.4\xspace}

\newcommand{\steerOffEdit}{76.6\xspace}

\newcommand{\ssSpRef}{86.5\xspace}
\newcommand{\ssSpArm}{97.0\xspace}
\newcommand{\ssSpLo}{6.8\xspace}
\newcommand{\ssSpHi}{14.8\xspace}
\newcommand{\fbFixSp}{34\xspace}

\newcommand{\fbDecSp}{1.4\xspace}
\newcommand{\fbFixPctSp}{74\xspace}
\newcommand{\ssSpD}{10.5\xspace}
\newcommand{\ssLtRef}{62.0\xspace}
\newcommand{\ssLtArm}{81.5\xspace}
\newcommand{\ssLtLo}{13.5\xspace}
\newcommand{\ssLtHi}{25.5\xspace}
\newcommand{\fbFixLt}{21\xspace}

\newcommand{\fbDecLt}{1.8\xspace}
\newcommand{\fbFixPctLt}{42\xspace}
\newcommand{\ssLtD}{19.5\xspace}
\newcommand{\steerInDS}{$-$1.0\xspace}
\newcommand{\steerInDSLo}{$-$4.0\xspace}
\newcommand{\steerInDSHi}{2.0\xspace}
\newcommand{\steerInRef}{93.0\xspace}
\newcommand{\steerInN}{100\xspace}
}{}

\title{Detecting Is Not Choosing: What a World Model Needs to Keep a VLA Policy Safe}

\author{
  Thanish Bolla\\
  Independent Researcher\\
  \texttt{thanish009.ai@gmail.com}
}

\begin{document}
\maketitle

\begin{abstract}
Vision-language-action (VLA) policies complete most benchmark tasks but often violate safety constraints. Since they are stochastic, one remedy is to sample several candidate action chunks and execute one that a world model predicts is safe. We ask what such a model must do. In simulation, a lookahead safety filter with the simulator as an exact world model raises the safety rate of three VLA policies on LIBERO and of \pizerofive on LIBERO-Plus by \gainMin to \gainLtPlus points. It also helps \pizerofive on three RoboCasa tasks. Learned critics trained on one policy's data detect dangerous states (pooled AUROC up to \critPooledMax) but rank candidates at one state close to chance (within-decision AUROC \critWithinMin--\critWithinMax), and the learned dynamics and video models we test do no better. When the simulator's state is perturbed, selection degrades long before detection. In the loop, a median position error of \radMedC\,mm removes \lostSpC\% of the gain on LIBERO-Spatial, partly because the error also corrupts the displacement check. On LIBERO-10 the loss is smaller, and the intervals overlap. The median does not predict the cost: a pixel reader with a \lpErrByst\,mm median error keeps \lpVsGaussAbs points less of the gain than Gaussian error of a similar median. A world model for safety should thus be judged by within-state ranking, and its state must be accurate in the tail and on handled objects, not only in the median.

\end{abstract}


\section{Introduction}\label{sec:intro}

Vision-language-action (VLA) policies now complete most tasks in standard manipulation benchmarks, but often unsafely. On LIBERO-10, \pizerofive succeeds in \succLtPiBase\% of our episodes, yet only \baseLtPi\% are free of violations such as pushing a non-target object, excessive contact force or tilting a carried object (Table~\ref{tab:guard}). Because the policies are stochastic, a simple remedy at test time is to sample several candidate action chunks and execute one that is predicted to be safe. This requires a world model. The question is what kind of model, and how accurate it must be.

Two families of approaches could make this prediction. Implicit approaches learn a critic or value function that scores candidates directly~\citep{nakamoto2024steering,kwok2025robomonkey,sun2026prevla}. Explicit approaches predict the future and evaluate it, as in latent safety filters~\citep{nakamura2025generalizing,seo2025uncertainty}, Cosmos Policy~\citep{kim2026cosmos} and learned predictive models combined with a geometric shield~\citep{zhong2026calibrated}. Both are usually judged in closed loop or by how well they separate safe from unsafe states pooled over a dataset. RoboMonkey's offline comparisons at one state use the demonstrated action, not safety, as the reference~\citep{kwok2025robomonkey}. A safety filter needs something narrower: at a single state, it must tell which of the policy's own candidates is safe.

We study this question with the most accurate world model available, the simulator itself. A lookahead safety filter rolls the policy's candidates forward from an exact copy of the simulator state and executes the one with the fewest predicted violations. For these candidates and this rule, no model can predict better, although another rule might be safer. Against it we place learned critics, dynamics and video models, and state estimators, and degrade the simulator's state by known amounts.

With the exact model, the filter raises safety in all \nGuardRows policy--benchmark pairs (\gainMin to \gainLtPlus points on LIBERO and LIBERO-Plus), for \pizerofive largely via fallbacks. The learned critics we train detect dangerous states no better than a critic that never sees the candidate, and within a state they rank candidates near chance, as do the explicit models we test. Selection degrades with state error long before detection, and real estimators lose more than their median error predicts. Our answer to the workshop's question: such a filter needs a world model that ranks actions at one state, and among the models and data we tested only the exact simulator does. Its state must also be accurate in the tail and on handled objects, where the median error says little. The test is therefore within-state ranking rather than pooled AUROC, and simulator labels make it cheap.

\section{Setup and method}\label{sec:method}

A VLA policy maps images, proprioception and an instruction to a chunk of $H{=}10$ actions (0.5\,s). The robot executes five and then queries the policy again at the next \emph{decision}. $K$ random seeds give $K$ candidate chunks. The first, the \emph{nominal chunk}, is what the policy would execute alone. We test \pizerofive~\citep{black2025pi05}, \pizerofast~\citep{pertsch2025fast} and SmolVLA~\citep{shukor2025smolvla} on LIBERO-Spatial and LIBERO-10~\citep{liu2023libero} (SmolVLA on LIBERO-Spatial only), and \pizerofive on LIBERO-Plus~\citep{fei2025liberoplus} and three RoboCasa tasks~\citep{nasiriany2024robocasa}, all in MuJoCo via robosuite~\citep{zhu2020robosuite}. Constraints and thresholds follow SafeVLA-Bench~\citep{fan2026safevlabench}, except for tilt on LIBERO-10. An episode is unsafe if a non-target object other than the task's receptacle moves more than 5\,mm or a contact force exceeds 200\,N. It is also unsafe if the arm collides with itself or a carried object tilts more than $15^\circ$ ($30^\circ$ on LIBERO-10). The \emph{safety rate} is the share of safe episodes.

The lookahead safety filter uses the simulator as its world model. At each decision it saves the simulator state, runs the nominal chunk for $H$ steps and restores the state exactly. If no constraint was violated, the robot executes the nominal chunk. Otherwise the filter simulates the other $K{-}1{=}15$ candidates and executes the one with the fewest predicted violations. At an \emph{infeasible} decision, where all candidates are unsafe, it also tries two short fallback motions, a lift and a lift away from the nearest non-target object, and runs one if it predicts fewer violations. With \pizerofive, the filter departs from the nominal chunk at \devSpPi\% of decisions on LIBERO-Spatial and \devLtPi\% on LIBERO-10.

Against it we test learned models of two kinds. An \emph{implicit} model, a learned safety critic, predicts the probability that a candidate violates a constraint within $H$ steps. We train six critics on simulator labels of all 16 candidates at \critTrainDec decisions. Five see the candidate with privileged object poses, policy features or images, and one sees only the state. An \emph{explicit} model predicts the future, on which the constraints are checked. We test a learned dynamics model, a latent world model on DINOv2 features~\citep{oquab2024dinov2} in the style of DINO-WM~\citep{zhou2025dinowm}, and Cosmos Policy's video predictions~\citep{kim2026cosmos}. We also run the simulator from the output of two state estimators, a convolutional pixel reader and FoundationPose~\citep{wen2024foundationpose}. To measure how exact the model must be, we perturb the state the lookahead reads: at each decision, every object the robot does not touch gets a horizontal offset $\mathcal{N}(0,\sigma^2)$ per axis, with $\sigma\in\{\sigA,\sigB,\sigC\}$\,mm. We report every error as the median horizontal distance to the true position (\radMedA, \radMedB and \radMedC\,mm here), the measure the estimators report (Appendix~\ref{app:details}).

Our measures separate detection from ranking. A \emph{contested} decision has both safe and unsafe candidates. \emph{Detection} is the AUROC of a prediction pooled over decisions, whereas \emph{within-decision AUROC} is computed over the 16 candidates of one contested decision and averaged, so it measures ranking at one state. \emph{Selection accuracy} is the share of contested decisions with an unsafe nominal chunk at which the chosen candidate is safe. The \emph{half-gain error} $\ehalf$ is the median state error at which this accuracy falls halfway to chance. In the loop, each filtered episode is paired with an unfiltered one sharing task, layout and sampling noise, with 95\% intervals from a bootstrap over (task, layout) clusters. The critic and state-error studies and two RoboCasa tasks were planned ahead.

\section{Results}\label{sec:results}

\begin{table}[t]
\centering
\footnotesize
\setlength{\tabcolsep}{2.5pt}
\begin{tabular}{@{}llrrrrr@{}}
\toprule
Policy & Benchmark & Pairs & Safety (\%) & $\Delta$Safety & Fixed\,/ & $\Delta$Success \\
& & & nominal$\,\rightarrow\,$guarded & [95\% CI] & broken & [95\% CI] \\
\midrule
\pizerofive & LIBERO-Spatial & 400 & 86.8 $\rightarrow$ 98.3 & +11.5 [7.8, 15.8] & 46\,/\,0 & $-$0.8 [$-$2.0, 0.3] \\
\pizerofive & LIBERO-10 & 200 & 70.5 $\rightarrow$ 94.0 & +23.5 [17.5, 29.5] & 50\,/\,3 & $-$5.0 [$-$8.0, $-$2.0] \\
\pizerofast & LIBERO-Spatial & 500 & 83.8 $\rightarrow$ 96.4 & +12.6 [9.6, 15.6] & 64\,/\,1 & $-$1.2 [$-$3.0, 0.4] \\
\pizerofast & LIBERO-10 & 500 & 40.4 $\rightarrow$ 63.4 & +23.0 [19.0, 27.2] & 122\,/\,7 & $-$1.4 [$-$4.2, 1.2] \\
SmolVLA & LIBERO-Spatial & 500 & 83.8 $\rightarrow$ 98.6 & +14.8 [11.8, 17.8] & 74\,/\,0 & +0.2 [$-$2.2, 2.6] \\
\addlinespace
\pizerofive & LIBERO-Plus-Spatial & 210 & 78.6 $\rightarrow$ 94.3 & +15.7 [11.0, 20.5] & 33\,/\,0 & $-$2.8\textsuperscript{\dag} \\
\pizerofive & LIBERO-Plus-10 & 210 & 65.7 $\rightarrow$ 91.4 & +25.7 [20.0, 31.9] & 55\,/\,1 & $-$2.9\textsuperscript{\dag} \\
\addlinespace
\pizerofive & RoboCasa (3 tasks) & 107 & 62.6 $\rightarrow$ 93.5 & +30.8 [23.4, 39.3] & 33\,/\,0 & +2.8 [0.0, 6.5] \\
\bottomrule
\end{tabular}

\vspace{2pt}
\caption{The filter with an exact simulator. Reference: the policy alone (LIBERO, LIBERO-Plus) or the policy with the lookahead running but not acting (RoboCasa; Appendix~\ref{app:details}). Fixed\,/\,broken: pairs made safe\,/\,unsafe. RoboCasa: the \nRC of \nRCall pairs whose first decision matches in both arms; task-stratified interval (by task cluster: [\gainRCClLo, \gainRCClHi]). \dag\,No interval. $\Delta$ from unrounded rates.}
\label{tab:guard}
\end{table}

The exact model makes every policy safer. The filter raises safety in all \nGuardRows policy--benchmark pairs, by \gainMin to \gainMax points (Table~\ref{tab:guard}), and makes at most \brokenMax episodes unsafe in each. The gain holds under LIBERO-Plus's perturbations and on RoboCasa, where most violations are tilts and forces. CoffeeSetupMug, chosen after an exploratory run, gives most of the RoboCasa gain, and on two of the three tasks \pizerofive rarely succeeds (Appendix~\ref{app:details}). Success changes most for \pizerofive on LIBERO-10, from \succLtPiBase\% to \succLtPiAbs\% ([\succLtPiLo, \succLtPiHi] points). All other point estimates are within three points.

\begin{figure}[t]
\centering
\begin{minipage}[t]{0.52\linewidth}\centering
\includegraphics[width=\linewidth]{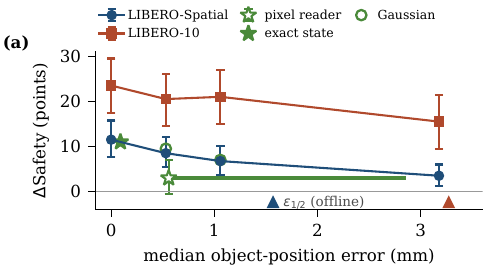}
\end{minipage}\hfill
\begin{minipage}[t]{0.46\linewidth}\centering
\includegraphics[width=\linewidth]{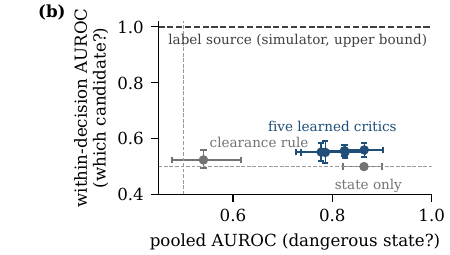}
\end{minipage}
\caption{(a) In-loop safety gain of the filter (\pizerofive) when the object positions it reads carry Gaussian error. Green (LIBERO-Spatial, one set of \lpN pairs): the filter on the pixel reader, from its median error on non-target objects to that on task objects, on Gaussian error (circles) and on the exact state (star, drawn at 0.1\,mm). Triangles: offline $\ehalf$ (displacement constraint). (b) Detection against ranking for the critics on held-out layouts (\critNcontested contested decisions). Five see the candidate, while the state-only critic does not and sits at 0.5 by construction, and the clearance rule ranks by the gripper's planned distance to non-target objects. Bars: 95\% intervals.}
\label{fig:main}
\end{figure}

\subsection{Can learned models tell the candidates apart?}
Every critic that sees the candidate detects dangerous states, with pooled AUROC \critPooledMin to \critPooledMax on held-out layouts (Fig.~\ref{fig:main}b). Yet the state-only critic scores \critPooledState, as high as any of them. The pooled score thus measures how dangerous the state is, not which action is safe, akin to objective mismatch~\citep{lambert2020objective}. Within one state, these five rank the 16 candidates barely better than chance, even the one given the full object state: within-decision AUROC is \critWithinMin to \critWithinMax, with no upper 95\% limit above \critWithinBestHi. A ranking loss does not change this ($\Delta$AUROC $=$ \critLossDelta).

The explicit learned models also rank near chance. A dynamics model trained on all candidates reaches a within-decision AUROC of \dynWithinCF (seeds \dynWithinCFMin--\dynWithinCFMax; \dynNcontested contested decisions, displacement constraint). The DINOv2 latent world model is near chance on both suites (\ltAucSp and \ltAucLt). Cosmos Policy predicts where disturbed objects go with a median error of \cwPred\,mm, against \cwStill\,mm for predicting no motion. At \cwSelN contested decisions with an unsafe nominal chunk, positions read from its predicted frames pick a safe candidate \cwSel of the time (chance \cwSelChance; true future frames: \cwSelTrue). As a selector, Cosmos's value head changes safety by \cosmosValDS points [\cosmosValDSLo, \cosmosValDSHi].

\subsection{How exact must the state be?}
Selection degrades with state error long before detection. Offline on LIBERO-Spatial, at a median error of \lvRadTwo\,mm the perturbed lookahead still detects decisions with an unsafe nominal chunk (AUROC \whenSpTwo), but selection has fallen from \selExactSp to \selSpTwo (chance \selChanceSp). On the displacement constraint, selection falls halfway to chance at $\ehalf = \halfRadSp$\,mm on LIBERO-Spatial (\halfRadSpLo--\halfRadSpHi\,mm, \halfNSp decisions) and at \halfRadLt\,mm on LIBERO-10. The DINOv2 latent reads current object positions to \ltErrSp\,mm, above $\ehalf$. Over all constraints on LIBERO-Spatial, even a median error of \evSpTinyRad\,mm lowers selection from \selExactSp to \evSpTiny [\evSpTinyLo, \evSpTinyHi] (chance \evChanceSp, \evSpTinyN decisions). Part of the exact model's advantage there needs an essentially exact state. On LIBERO-10, selection is still \stateLtEventEight at \lvRadEight\,mm.

In the loop (Fig.~\ref{fig:main}a), a median error of \radMedC\,mm removes \lostSpC\% of the gain on LIBERO-Spatial but only \lostLtC\% on LIBERO-10, where every level's interval overlaps that of the exact state. Part of this loss is blindness: at large errors the copy sees objects past the 5\,mm limit and cannot flag new violations. Through the predicted dynamics alone, the loss is \lostSpCdyn\% on LIBERO-Spatial and \lostLtCdyn\% on LIBERO-10 (Appendix~\ref{app:details}).

Real state estimators lose more than their median error predicts. The pixel reader reads non-target objects to a median of \lpErrByst\,mm, well under $\ehalf$, but to \lpErrBystPn\,mm at the 90th percentile. On the same \lpN pairs it keeps \lpVsGaussAbs points less of the gain than Gaussian error of similar median (\radMedA\,mm; [\lpVsGaussLo, \lpVsGaussHi]). Its error on task objects is larger (median \lpErrTarg\,mm, 90th percentile \lpErrTargPn\,mm), and at \lpBelief\% of decisions it makes the copy see a constraint as already violated, silencing the filter. FoundationPose~\citep{wen2024foundationpose} shows the same gap offline: on LIBERO-Spatial its median errors on both kinds of object are under $\ehalf$, yet it keeps only \fpRescueTrSp [\fpRescueTrSpLo, \fpRescueTrSpHi] of the \fpRescueNSp rescues made with true poses. On LIBERO-10, where the curve is flatter, it keeps \fpRescueTrLt [\fpRescueTrLtLo, \fpRescueTrLtHi] of \fpRescueNLt, although it loses track of moved task objects (Appendix~\ref{app:details}). In both estimators the median does not predict the cost.

\section{Related Work}\label{sec:related}

Other selectors train a verifier on decision-critical states~\citep{park2026diver} or steer sampling noise~\citep{wagenmaker2025steering,chen2026inspo}. VLAPS and V-VLAPS search in a simulator assumed accurate~\citep{neary2025vlaps,ren2026vvlaps}. We ask how accurate it must be.

Safety can also be trained into a VLA~\citep{zhang2025safevla,tayal2026shieldvla} or imposed by barrier functions or guidance fields around detected regions~\citep{hu2025vlsa,park2026attention,agarwal2026multilink,sinaei2026safe,song2026omniguide}. In the terms of \citet{hsu2024safety}, these are geometric safety filters. None predicts contact forces or pushed objects, which our filter simulates.

Learned safety models are usually judged in closed loop or by detection pooled over states: safe-RL critics~\citep{srinivasan2020learning,thananjeyan2021recovery,bharadhwaj2021conservative}, latent safety filters~\citep{nakamura2025generalizing,seo2025uncertainty}, outcome predictors for policy steering~\citep{wu2025foresight} and failure detectors~\citep{zhang2026foresight}. \citet{zhong2026calibrated} score VLA candidates with a learned model and report pooled AUROC, leaving reranking untested (their Sec.~5.1). We measure that ranking directly.

\section{Discussion}\label{sec:discussion}

At \ceilSp\% of decisions with an unsafe nominal chunk on LIBERO-Spatial (LIBERO-10: \ceilLt\%), none of 64 candidates is safe (Appendix~\ref{app:ceiling}), so even an exact model needs a fallback. For \pizerofive, fallbacks run in \fbFixPctSp\% of fixed LIBERO-Spatial episodes and \fbFixPctLt\% on LIBERO-10, though at only \fbDecSp--\fbDecLt\% of decisions. But 41 of \fixedSpPi and 48 of \fixedLtPi fixed episodes also switch to another of the policy's candidates, and only 5 and 2 are fixed by fallbacks alone. Yet fallbacks on the nominal chunk alone, without ranking, recover most of the gain (Limitations).

The learned-model results hold for the models and data we tested: critics trained on one policy, with only \critTrainContested of \critTrainDec training decisions contested. The best critic still picks a safe candidate \critSelP of the time [\critSelPLo, \critSelPHi] at contested decisions where the nominal chunk is unsafe (chance \critSelChance), but pooled detection does not show whether a model can rank, and ours do not rank well enough to replace the exact simulator.

\section*{Limitations}\label{sec:limitations}

\paragraph{Fallbacks.} A fallback runs only when every candidate is predicted unsafe and the lookahead predicts fewer violations for it. We also ran each part alone on the paper's \pizerofive episodes. Ranking without fallbacks falls short of the full filter by 5.8 points [3.0, 9.0] on LIBERO-Spatial and 9.5 [5.0, 14.5] on LIBERO-10. Trying the fallbacks whenever the nominal chunk is predicted unsafe, with no ranking, comes within 0.8 [$-1.0$, 2.5] and 3.5 [$-0.5$, 8.0] points of it, but lowers success on LIBERO-10 by 7.0 points, against 5.0 for the full filter. For \pizerofive, most of the gain thus needs only an exact comparison at one state between the nominal chunk and up to two fallbacks, and ranking the policy's other candidates adds little once fallbacks exist. We have not tested this for the other policies. Nor did we run a geometric rule in the loop. Offline, a hand-written clearance rule ranks no better than the critics (Fig.~\ref{fig:main}b).

\paragraph{Training data for the learned models.} Contested decisions, with both safe and unsafe candidates, are the states where ranking can be learned, and our critics see few of them, all from one policy. In a later check with retrained critics, within-decision ranking improved slightly from a quarter of the contested training decisions to all of them, while the rate of picking a safe candidate did not change detectably. More distinct contested states may therefore rank better, whereas re-weighting the existing ones (up-weighting them, or training on them alone) did not measurably help. Methods designed for such states~\citep{park2026diver} remain untested here.

\paragraph{Benchmarks and costs.} All results are in simulation, and the filter needs an exactly restorable state, which failed on two PhysX benchmarks (ManiSkill3~\citep{tao2024maniskill3}, SimplerEnv~\citep{li2024simpler}). The learned-model analysis rests on LIBERO, whose narrow training and test conditions VLA policies are known to overfit~\citep{fei2025liberoplus}. On LIBERO-Plus we test only the exact filter, so the candidate diversity and failure patterns we rank may be specific to LIBERO. RoboCasa covers three tasks, CoffeeSetupMug gives most of its gain, and \pizerofive rarely succeeds on the other two. \pizerofast and SmolVLA run on LIBERO only, and $\ehalf$ rests on few decisions. On LIBERO-10 with all constraints, a 10\% friction error lowers offline selection more (\physLtFricMinus at $-10\%$, \physLtFricPlus at $+10\%$) than a \lvRadEight{}\,mm position error (\stateLtEventEight). The filter trades some success for safety: for \pizerofive on LIBERO-10, success falls from \succLtPiBase\% to \succLtPiAbs\%, although episodes both safe and successful rise by \ssSpD points on LIBERO-Spatial and \ssLtD on LIBERO-10 (Appendix~\ref{app:details}). The filter simulates the other candidates only when the nominal chunk is predicted unsafe, so it simulates 1.5 to 1.8 chunks per decision on average (at most 18). Measured with six rollout workers on one 8-vCPU machine, the lookahead takes 0.3 to 0.4\,s on average where the nominal chunk is predicted safe and 5 to 8\,s (at most 12.4\,s) where it is not, slower than the 0.25\,s the robot executes per decision, so the filter does not run in real time on this hardware.

\section*{AI usage statement}

AI coding assistants were used to help write the code for this project. The author takes full responsibility for the content.

\acknowledgments{We thank the two anonymous reviewers and the program chairs for their feedback.}

\clearpage
\bibliography{refs}

\clearpage
\appendix
\section{Experimental details}\label{app:details}

\paragraph{Filter.}
Among unsafe candidates, ties in the number of predicted violations go to the smaller violation depth, then to the lower index. The fallback motions hold the gripper command and lift the end effector by about 2\,cm over the five executed steps; the second also moves it about 1.2\,cm away from the nearest non-target object. The lookahead always predicts the tilt constraint at $15^\circ$, so on LIBERO-10 it is stricter than the score. For \pizerofast and SmolVLA the 15 extra candidates are drawn only after the nominal chunk is predicted unsafe, which leaves the nominal chunk unchanged. For \pizerofive, \fbFixSp of the \fixedSpPi fixed episodes on LIBERO-Spatial and \fbFixLt of \fixedLtPi on LIBERO-10 ran at least one fallback. The share of episodes both safe and successful rises from \ssSpRef\% to \ssSpArm\% on LIBERO-Spatial ([\ssSpLo, \ssSpHi] points) and from \ssLtRef\% to \ssLtArm\% on LIBERO-10 ([\ssLtLo, \ssLtHi]).

\paragraph{Pairing and references.}
For \pizerofive we use layouts 30--49 of every task, with two sampling-noise seeds on LIBERO-Spatial (\nSpPi pairs) and one on LIBERO-10 (\nLtPi). For \pizerofast and SmolVLA we use all 50 initial layouts of each task (\nSpFast pairs per suite). SmolVLA is not run on LIBERO-10, where its success was below the bar we set in advance. On LIBERO the reference is the policy alone. Running the lookahead while always executing the nominal chunk does not change the simulator state. Over \seN paired \pizerofive episodes, safety differed in \seSafeDisc and success in \seSuccDisc; where logged trajectories differ, the first difference is in a rendered image. On RoboCasa the reference runs the lookahead and always executes the nominal chunk. CoffeeSetupMug was chosen after an exploratory labelling run and the other two tasks by a screen written in advance; after fixing a rendering bug we reran the unchanged protocol. The pooled gain is not uniform: CoffeeSetupMug gains \rcCoffeeGain points [\rcCoffeeLo, \rcCoffeeHi] on \rcCoffeeN pairs (\rcCoffeeFix of the \fixedRC rescues), with success \rcCoffeeSuccBase\% unfiltered and \rcCoffeeSuccGuard\% filtered. PickPlaceSinkToCounter gains \rcSinkGain [\rcSinkLo, \rcSinkHi] on \rcSinkN pairs at \rcSinkSuccBase\% success, and CloseBlenderLid gains \rcBlenderGain [\rcBlenderLo, \rcBlenderHi] on \rcBlenderN pairs, with success \rcBlenderSuccBase\% and \rcBlenderSuccGuard\%. Rates have Wilson intervals. In-loop differences use a bootstrap over (task, layout) clusters with 2{,}000 resamples; offline selection accuracies and $\ehalf$ use one over episodes with 1{,}000.

\paragraph{State perturbation.}
Offsets are drawn independently at every decision for every free object that is not in contact with the robot. The perturbed positions enter every quantity of the check, including the displacement of non-target objects from their start positions. A constraint that the perturbed copy already sees as violated at the decision raises no new event. In a second variant the displacement reference moves with the offset, so the error enters only through the predicted dynamics. At the largest level this variant keeps \keptSpCdyn\% of the gain on LIBERO-Spatial (\keptLtCdyn\% on LIBERO-10), \dynMinusStaticSp points [\dynMinusStaticSpLo, \dynMinusStaticSpHi] more than the main variant. Offline, the same perturbation is applied to saved decisions of \pizerofive with all 16 candidates labelled (\halfNSp contested decisions with an unsafe nominal chunk on LIBERO-Spatial, \halfNLt on LIBERO-10, for the displacement constraint). Offline levels are $\sigma\in\{0.01, 0.1, 0.25, 0.5, 1, 2, 4, 8\}$\,mm per axis, and $\ehalf$ is interpolated linearly between them; over all constraints on LIBERO-Spatial it is \evHalfRadSp\,mm [\evHalfRadSpLo, \evHalfRadSpHi]. Chance is the mean share of safe candidates at the same decisions, so it differs between decision sets. The interval for $\ehalf$ on LIBERO-10 is \halfRadLtLo--\halfRadLtHi\,mm, and chance over all constraints there is \evChanceLt.

\paragraph{Learned models.}
The critics are trained on initial layouts 0--29, selected on 30--39 and tested on 40--49 (\critNcontested contested decisions, \critNepisodes episodes). Only \critTrainContested of the \critTrainDec training decisions are contested, so a critic sees few states where the choice matters. At the \critNdanger contested test decisions with an unsafe nominal chunk, the best critic picks a safe candidate \critSelP of the time [\critSelPLo, \critSelPHi] (chance \critSelChance). The image critics use a fine-tuned ResNet-18 or frozen SigLIP~2~\citep{tschannen2025siglip2}. One critic is trained with a within-decision ranking loss and one with binary cross-entropy. The learned dynamics model is an MLP that predicts object and end-effector positions over the lookahead from the state and the candidate chunk. It is trained either on the executed steps or on all 16 candidates. Even selecting on the true simulated displacement picks a safe candidate at only \cwSelSim of the \cwSelN decisions used for Cosmos Policy, because a candidate that moves no object can still violate the force or tilt constraint. FoundationPose's median error on LIBERO-Spatial is \fpTrSpByst\,mm on non-target and \fpTrSpTarg\,mm on task objects; on LIBERO-10, its 90th-percentile error on moved task objects is \fpTrLtTargPn\,mm. The pixel reader is trained on rendered frames of layouts 30--37 and tested on 40--49.

\section{Selection ceiling}\label{app:ceiling}
For \pizerofive on LIBERO-Spatial and LIBERO-10, going from 16 to 64 candidates removes only a quarter of the infeasible decisions (ratio \ceilRatioSp [\ceilRatioSpLo, \ceilRatioSpHi] on LIBERO-Spatial). At \restN decisions where all 16 candidates are unsafe, fresh sampling noise frees \restNoise\%, a zero-velocity pause \restZeroVel\%, letting the objects settle \restSettle\%, holding the arm still \restHold\% and moving the arm first \restArm\%. Small structured edits of the nominal chunk free \steerOffEdit\% of the \steerOffN decisions at which 64 samples fail, against \steerOffSamples\% for 64 more samples. In the loop, however, using these edits in place of the fallback does not raise the safety rate (\steerInDS points [\steerInDSLo, \steerInDSHi], \steerInN pairs, with the filter already at \steerInRef\%).

\end{document}